\documentclass[letterpaper]{article} 
\usepackage{aaai24}  
\usepackage{times}  
\usepackage{helvet}  
\usepackage{courier}  
\usepackage[hyphens]{url}  
\usepackage{graphicx} 
\usepackage{natbib}  
\usepackage{caption} 
\usepackage{algorithm}
\usepackage{algorithmic}

\usepackage{booktabs}       
\usepackage{amsfonts}       
\usepackage{amsmath}
\usepackage{xcolor}         
\usepackage{xspace}
\usepackage{multirow}
\usepackage{subcaption} 
\usepackage{comment}
\usepackage{newfloat}
\usepackage{listings}
\DeclareCaptionStyle{ruled}{labelfont=normalfont,labelsep=colon,strut=off} 
\floatstyle{ruled}
\newfloat{listing}{tb}{lst}{}
\floatname{listing}{Listing}
\newcommand{\xvec}{{\bf x}}
\newcommand{\yvec}{{\bf y}}
\newcommand{\bvec}{{\bf b}}

\newcommand{\ReLU}{\ensuremath{\mathit{ReLU}}\xspace}

\newcommand{\Clip}{\ensuremath{\mathit{Clip}}\xspace}

\newcommand{\Round}{\ensuremath{\mathit{Round}}\xspace}

\title{Towards Efficient Verification of Quantized Neural Networks}
\author{
    Pei Huang\textsuperscript{\rm 1},
    Haoze Wu\textsuperscript{\rm 1},
    Yuting Yang\textsuperscript{\rm 2},
    Ieva Daukantas\textsuperscript{\rm 3},\\
    Min Wu\textsuperscript{\rm 1},
    Yedi Zhang\textsuperscript{\rm 4} and 
    Clark Barrett\textsuperscript{\rm 1}\thanks{Corresponding Author.}
}
\affiliations{
    \textsuperscript{\rm 1}Stanford University, Stanford, USA\\
    \textsuperscript{\rm 2}Institute of Computing Technology, Chinese Academy of Sciences, Beijing, China\\
    \textsuperscript{\rm 3} IT University of Copenhagen, Denmark\\
    \textsuperscript{\rm 4} National University of Singapore, Singapore\\
    \{huangpei,haozewu,minwu\}@stanford.edu, yangyuting@ict.ac.cn\\ daukantas@itu.dk, yd.zhang@nus.edu.sg, barrett@cs.stanford.edu
    
}

\usepackage{bibentry}

\begin{document}

\maketitle

\begin{abstract}
\emph{Quantization} replaces floating point arithmetic with integer arithmetic in deep neural network models, providing more efficient on-device inference with less power and memory.
In this work, we propose a framework for formally \emph{verifying} properties of quantized neural networks.
Our baseline technique is based on integer linear programming which guarantees both \emph{soundness} and \emph{completeness}. We then show how efficiency can be improved by utilizing gradient-based heuristic search methods and also bound-propagation techniques.
We evaluate our approach on perception networks quantized with PyTorch. Our results show that we can verify quantized networks with better scalability and efficiency than the previous state of the art.
\end{abstract}

\section{Introduction}

In recent years, deep neural networks (DNNs) \cite{dl} have demonstrated tremendous capabilities across a wide range of tasks \cite{vgg,bert,visionTF}. However, DNNs have also shown various security and safety issues, e.g., vulnerability to input perturbations \cite{GoodfellowSS14,epsilon,FPP,yang}. Such issues must be addressed before DNNs can be used in safety-critical scenarios such as autonomous driving \cite{XuGYD17} and medical diagnostics \cite{CiresanGGS12}. Formal verification is an established technique which applies mathematical reasoning to ensure the correct behavior of safety-critical systems, and several approaches for applying formal methods to DNNs have been investigated~\cite{HuangKWW17,LechnerZCH22}.

Our focus is the verification of quantized neural networks (QNNs).  Quantization replaces inputs and parameters represented as 32/64-bit floating point numbers with a lower bit-width fixed point (e.g., 8-bits) representation~\cite{JacobKCZTHAK18,HanMD15}. QNNs can greatly reduce both memory requirements and computational costs while maintaining competitive accuracy. As a result, they are increasingly being used in embedded applications, including safety-critical applications such as autonomous driving. For instance, 8-bit quantized DNNs have been applied in Tesla's Full Self-Driving Chip (previously Autopilot Hardware 3.0)~\cite{henzinger2021,TeslaWeb}. With the increasing popularization and use of QNNs, it is urgent to develop efficient and effective verification techniques for them.

In this work, we propose an efficient verification framework for QNNs with three components, offering different trade-offs between scalability and precision. The baseline approach models neural networks and formal properties as integer linear programming (ILP) problems. ILP is an exact method in the sense that it guarantees both \emph{soundness} (if it reports the system is safe, then it really is safe) and \emph{completeness} (if the system really is safe, then it will report that it is safe). Unlike previous work, which focuses on 
simple models of quantized neural networks, ours is the first formal approach that precisely captures the quantization scheme used in popular deep learning frameworks such as PyTorch/TensorFlow.

Our ILP approach is precise but may encounter scalability issues on larger QNNs. To address this, we also propose a gradient-based method for finding counterexamples.  We use a rewriting trick
for the non-differentiable round operation, which enables the backward process to cross
through the round operation and gives us the desired gradient information.
If this method finds a counterexample, then we immediately know that the property does not hold, without having to invoke the ILP solver.

\begin{figure*}
    \begin{minipage}[b]{0.28\textwidth}
        \includegraphics[width=\textwidth]{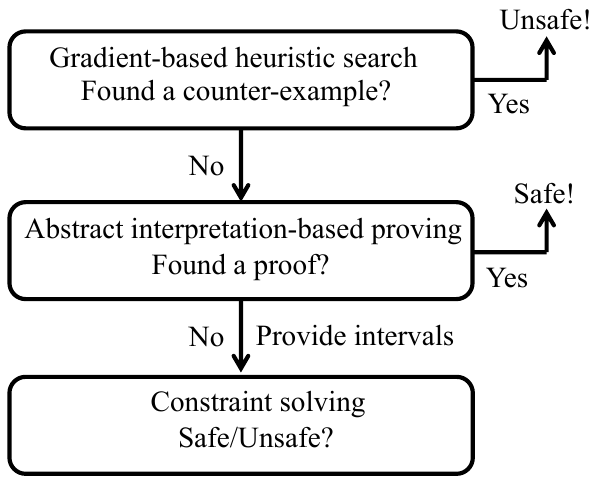}
        \caption{The main verification framework for QNN.}
        \label{fig:verifram}
    \end{minipage}%
    \hfill
\begin{minipage}[b]{0.7\textwidth}
\includegraphics[width=\linewidth]{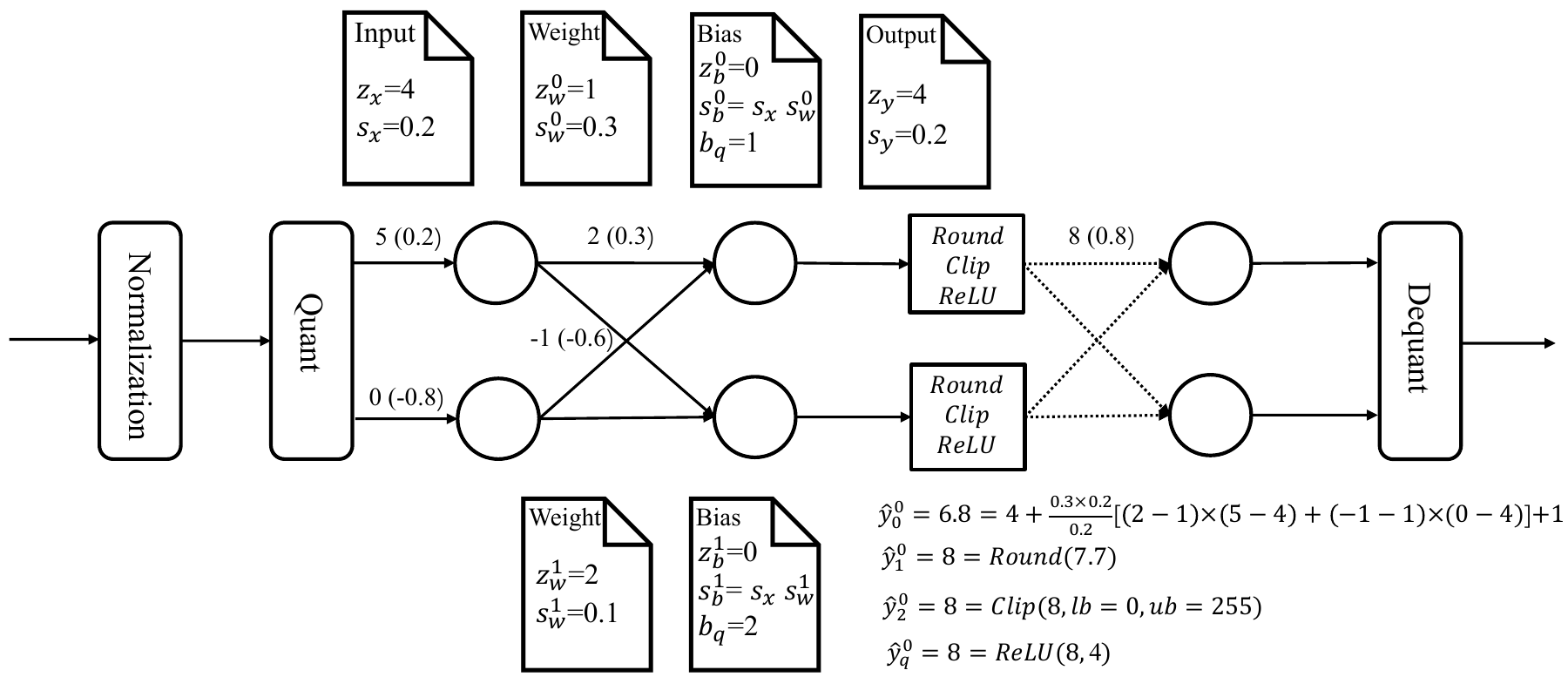}
\caption{Quantized neural network. A number in the form of $I(F)$ means that ``$I$'' is its integer representation and  ``$F$'' is its corresponding fixed-point representation.}
\label{fig:qnn}
    \end{minipage}%
\end{figure*}

The third component of the framework lies in between the first two. It relies on abstract interpretation-based reasoning to do an incomplete but formal analysis.  We extend existing abstract interpretation-based interval analysis methods to support the semantics of ``round" and ``clip" operations in quantized neural networks. In particular, for the clip operation, we reduce it to a gadget built from two \ReLU units.
If the abstract interpretation approach succeeds, we know the property holds. Otherwise, the result of the analysis can be used to reduce the runtime of the ILP-based complete method. The overall framework is depicted in~Fig. \ref{fig:verifram}. 



Based on our framework, we realize an \textbf{E}fficient \textbf{Q}NN \textbf{V}erification system named EQV. We use EQV to verify the robustness of QNNs against bounded
input perturbations. Our experimental results show that EQV can scale to networks that are more than twice as large as the largest networks handled by previous approaches. We also show that, compared to the baseline ILP technique, EQV is up to 100 $\times$ more efficient for some cases. Our contributions can be summarized as the following:
(1) We provide a ILP-based exact verification approach for the QNNs which first precisely captures the quantization scheme used in current popular deep learning frameworks;
(2) We extend existing abstract interpretation-based interval analysis methods to support QNNs;
(3) We design a rewriting trick for the non-differentiable round operation, which enables gradient-based analysis of QNNs;
(4) We implement our approach in a tool, EQV, and demonstrate that it can scale to networks that are twice the size of the largest analyzed by the current state-of-the-art methods, and up to 100 $\times$ faster than the baseline ILP method.

\section{Background and Related Work}

Formal DNN verification checks whether a DNN satisfies a property such as the absence of adversarial examples in a given perturbation space. The property is usually depicted by a formal specification, and verifiers aim to provide either a proof of the validity of this property or a counterexample. Researchers have
developed a range of verification techniques, mostly for real-valued \ReLU networks. Exact methods (i.e., sound and complete) can always, in theory, answer whether a property holds or not in any situation. Typical exact methods formalize the verification problem as a Satisfiability Modulo Theories (SMT) problem \cite{KatzBDJK17,Ehlers17,HuangKWW17,JiaH00MZ23} or a Mixed Integer Linear Programming (MILP) problem \cite{ChengNR17,FischettiJ18,DuttaJST18}, but their scalability is limited as the problem is NP-hard~\cite{KatzBDJK17}. Another typical approach is to use a method that only guarantees soundness, i.e., to improve the scalability at the cost of completeness. Abstract interpretation is one such approach.  It overapproximates the behavior of the neural network with the hope that the property can still be shown to hold ~\cite{WongK18,WengZCSHDBD18,GehrMDTCV18,ZhangWCHD18,RaghunathanSL18,MirmanGV18,SinghGPV19}. Finally, heuristic approaches can be used to search for counterexamples. These techniques are neither sound nor complete but can be effective in practice~\cite{GoodfellowSS14,YangDPSZ22,SerbanPV20}.

Existing work on DNN verification typically focuses on networks whose parameters are real or floating point numbers. In contrast, relatively little prior work addresses the verification of QNNs.  QNN verification presents additional challenges due to the difficulty of modeling quantization schemes.  And some evidence suggests that it may also be more computationally challenging.
For example, Jia et al. \cite{JiaR20} point out that the verification of binarized neural networks (which can be regarded as 1-bit quantized neural networks) has exhibited even worse scalability than real-valued neural network verification. 

In the last two years, some work has started to focus on the verification of QNNs. Henzinger et al. \cite{henzinger2021} provide an SMT-based method to encode the problem as a formula in the SMT theory of bit-vectors. Mistry et al. \cite{MistrySB22} and hang et al. \cite{ZhangZCSZCS22} propose using MILP and ILP to model the QNN verification problem. All of these methods pioneer new directions for QNNs but are applied only to small models using simple quantization schemes.  None of them can directly support the sophisticated quantization schemes used in real deep learning frameworks.

\section{Preliminaries}
\label{Pre}
The quantization operation is a mapping from a real number $\gamma$ to an integer $q$ of the form
\begin{equation}\label{quan}
  \text{Quant:}\ \  q=Round(\frac{\gamma}{s}+z), \ \text{De-quant:}\ \   \gamma=s(q-z),
\end{equation}
for some constants $s$ and $z$. Equation \ref{quan} is the quantization scheme, and the constants $s$ and $z$ are quantization parameters.
The constant $s$ (for “scale”) is an arbitrary real number. The constant $z$ (for “zero point”) is the integer corresponding to the quantized value $q$ when $\gamma=0$. In practice, $q$ is represented using a fixed number of bits.  For example, in 8-bit quantization, $q$ is an 8-bit integer.  Note that in general, $q$ may not fit within the number of bits provided, in which case the closest representable value is used.  

One of the most important operations when doing forward inference in DNNs is matrix multiplication. Suppose we have three $N \times N$ matrices of real numbers, where the third matrix is equal to the product of the first two matrices. Denote the entries of these 3 matrices as $r_{\alpha}^{(i,j)}$, where $\alpha\in\{1,2,3\}$ and $0\leq i,j \leq N-1$. Their quantization parameters are $(s_{\alpha},z_{\alpha})$ (in general, different quantization parameters may be used for different neurons in a DNN). We use $q_{\alpha}^{(i,j)}$ to denote the quantized entries of these 3 matrices. Based on the quantization scheme $r_{\alpha}^{(i,j)}=s_{\alpha}(q_{\alpha}^{(i,j)}-z_{\alpha})$ and the definition of matrix multiplication, we have
\begin{equation}
    s_{3}(q_{3}^{(i,j)}-z_3)=\sum_{k=0}^{N-1} s_{1}(q_{1}^{(i,k)}-z_1)s_{2}(q_2^{(k,j)}-z_{2}),
\end{equation}
which can be rewritten as
\begin{equation}
    q_{3}^{(i,j)}=z_{3}+\frac{s_{1}s_{2}}{s_{3}}\sum_{k=0}^{N-1} (q_{1}^{(i,k)}-z_1)(q_2^{(k,j)}-z_{2}).
\end{equation}

Suppose $\yvec:=\ReLU(W\xvec+\bvec)$ is the function describing the transformation performed in a single layer of a DNN. Its quantized version $\yvec_q:=g(\xvec_q, W_q,\bvec_q)$ can be described by the series of calculations shown in Equation~(\ref{qnnprocess}), where $W_q$, $\bvec_q$, $\xvec_q$ and $\yvec_q$ are the quantized versions of the weight matrix $W$, bias vector $\bvec$, input vector $\xvec$, and output vector $\yvec$, respectively. Let $z_\xvec$ and $z_\yvec$ be the zero points of $\xvec$ and $\yvec$ respectively. As the zero points of the weights corresponding to each output neuron may be different, we use $z_{w}^{j}$ to denote the zero point of the weights corresponding to the $j$-th neuron. Similarly, $s_{w}^{j}$, $s_{\xvec}$, and $s_{\yvec}$ are the scales for the weight matrix, input, and output, respectively. The $\ReLU$ function in the quantized network can be represented as the maximum of the input and the zero point. The calculation for $\yvec_q:=g(\xvec_q, W_q,\bvec_q)$ can then be written as:

\begin{equation}\label{qnnprocess}
\begin{aligned}
  \text{(i)}&\ \hat{y}_{0}^{j} :=z_{y}+\frac{s_{w}^{j}s_{x}}{s_{y}}\sum_{i}(w_{q}^{(i,j)}-z_{w}^{j})(x_{q}^{i}-z_{x})+b_{q}^{j} \\
  \text{(ii)}&\ \hat{y}_{1}^{j}:=\Round(\hat{y}_{0}^{j})\\
  \text{(iii)}&\ \hat{y}_{2}^{j}:=\Clip(\hat{y}_{1}^{j},lb,ub)\\
  \text{(iv)}&\ y_{q}^{j}:=\max(\hat{y}_{2}^{j},z_y)
\end{aligned}
\end{equation}
where $lb$ and $ub$ are the smallest and largest values, respectively, that can be represented by our quantized integer type, e.g., for an 8-bit unsigned type, $[lb,ub]$=$[0,255]$. The \Clip function returns the value within $[lb,ub]$ closest to its input.  In Pytorch, weights are usually quantized as signed integers while the inputs and outputs of each layer are quantized as unsigned integers. The quantization parameters (i.e., zero points and scales) are computed offline and determined at the time of quantization. In the inference phase, they are constants. Fig.~\ref{fig:qnn} shows a QNN performing an example computation.

The property utilized in this paper for testing the verification efficiency is robustness. Let $f:\mathbb{D}^n \rightarrow \mathbb{O}^m$ be a neural network classifier, where, for a given input $x \in \mathbb{D}^n$,    $f(x)=\{o_0(x),o_1(x),..., o_{m-1}(x)\}\in \mathbb{O}^m$ represents the confidence values for $m$ classification labels. In general $\mathbb{D}$ and $\mathbb{O}$ are sets of real numbers, and for quantized neural networks they are sets of integers corresponding to the quantization type. The prediction of $x$ is given as $F(x)=\mathop{\arg\max}_{0\leq i \leq m-1} o_i(x)$, and the label space is denoted as $\mathcal{Y}$. The robustness property can be depicted as: given a test point $x_*$ with label $l_*$, a neural network is locally robust at point $x_*$ with respect to a perturbation radius $r$ if the following formula holds:
	\begin{equation}\label{rob}
		\forall x \  (x \in B_\infty(x_*, r) \rightarrow F(x)=l_*)
	\end{equation}
	where $B_\infty(x_*, r)=\{x\ | \left\| x-x_*\right\|_\infty \leq r \}$ is the perturbation space around $x_*$ bounded by an $\ell_\infty$-norm ball of radius $r$. The goal of the verifier is to answer whether Equation~(\ref{rob}) holds.

\section{ILP Modeling}

In this section, we introduce an ILP formulation for the QNN robustness verification problem. Compared with previous work on QNN verification~\cite{MistrySB22,ZhangZCSZCS22}, the main difference is that our encoding correctly models quantization schemes used in mainstream deep learning frameworks (e.g., PyTorch). In addition, unlike \cite{MistrySB22}, we avoid using floating point variables, as in our experience, it is easier to solve ILP problems than to solve MILP problems. And in contrast to \cite{ZhangZCSZCS22}, we avoid piecewise constraints which introduce many redundant variables. 

In this paper, we use a symbol with a dot (``$\cdot$'') to denote a variable in our ILP model corresponding to an input to output from some layer of the DNN, e.g.,  variable $\dot{y}$.

We show how to encode each step in calculation~\eqref{qnnprocess}.
For step (i), we use the variable $\dot{x}_{q}^{i}$ for the $i$-th component of the input and an auxiliary variable $\hat{y}_{0}^{j}$ to denote the result. Note that $\hat{y}_{0}^{j}$ is a temporary variable and is not of integer type. The introduction of this symbol is for the sake of convenience, and we show how to eliminate it below.

\begin{equation}\label{encoding_sum}
    \hat{y}_{0}^{j} =z_{y}+\frac{s_{w}^{j}s_{x}}{s_{y}}\sum_{i}(w_{q}^{(i,j)}-z_{w}^{j})(\dot{x}_{q}^{i}-z_{x})+b_{q}^{j}
\end{equation}

For step (ii), $\hat{y}_{1}^{j}:=\Round(\hat{y}_{0}^{j})$ can be encoded by the following two constraints:
\begin{equation}\label{round_encoding}
\left\{
\begin{array}{l}
\dot{\hat{y}}_{1}^{j}-\hat{y}_{0}^{j} \leq 0.5 \\
\hat{y}_{0}^{j}-\dot{\hat{y}}_{1}^{j} \leq 0.5-\varepsilon,
\end{array}
\right.
\end{equation}
where a small constant $\varepsilon$ is used to realize the ``$<$'' operator (in ILP solvers, this operator is usually not supported directly). Since the result of the sum in Equation~\eqref{encoding_sum} is always an integer, we can find a proper value for $\varepsilon$ based on the factor $s_{w}^{j}s_{x}/s_{y}$ which guarantees the correctness of the encoding.

We now eliminate the temporary variable $\hat{y}_{0}^{j}$ by combining constraints (\ref{encoding_sum}) and (\ref{round_encoding}):
\begin{equation}\label{affine_encoding}
\left\{
\begin{array}{l}
\dot{\hat{y}}_{1}^{j}-    z_{y}-\frac{s_{w}^{j}s_{x}}{s_{y}}\sum_{i}(w_{q}^{(i,j)}-z_{w}^{j})(\dot{x}_{q}^{i}-z_{x})-b_{q}^{j} \leq 0.5 \\
    z_{y}+\frac{s_{w}^{j}s_{x}}{s_{y}}\sum_{i}(w_{q}^{(i,j)}-z_{w}^{j})(\dot{x}_{q}^{i}-z_{x})+b_{q}^{j}-\dot{\hat{y}}_{1}^{j} \leq 0.5-\varepsilon
\end{array}
\right.
\end{equation}

Let $Encode$\_$max(z, x, y)$ denote the ILP encoding of $z=\max(x,y)$ which can be realized with big M method~\cite{ChengNR17}:

\begin{equation}
Encode\_max(z,x,y)=\left\{
\begin{array}{l}
b_{x}+b_{y}=1\\
x-z \leq 0\\
y-z \leq 0\\
x-z+Mb_{y} \geq 0\\
y-z+Mb_{x} \geq 0\\
y-x+Mb_{x} \geq 0
\end{array}
\right.
\end{equation}
where $M$ is a very large positive constant and $b_{x}$, $b_{y}$ are fresh 0-1 type integer variables. It is worth noting that we use this same encoding even if one of $x$ and $y$ is a constant value.

For step (iii) $\hat{y}_{2}^{j}:=\Clip(\hat{y}_{1}^{j},lb,ub)$, the constraints are:
 
\begin{equation}\label{clip_encoding}
 Encode\_max(\dot{\hat{y}}_{max}^{j}, \dot{\hat{y}}_{1}^{j}, lb)\ \bigcup \ Encode\_min(\dot{\hat{y}}_{2}^{j}, \dot{\hat{y}}_{max}^{j}, ub)
\end{equation}
 where $\dot{\hat{y}}_{max}^{j}$ is a fresh auxiliary variable. Finally, step (iv) can be directly written as $Encode\_max(\dot{\hat{y}}_{q}^{j},\dot{\hat{y}}_{2}^{j},z_y)$.

 In our ILP model, the input variables represent the values of the input after input quantization. So the upper and lower bounds of the perturbation space also need to be quantized when representing the input constraint.

\subsection{Encoding for Typical Fusion Layers}
In order to reduce the amount of computation required for a quantized neural network during inference, certain layers are fused by the quantization process so that one kernel call does the computation for several neural network layers. We can use the same approach to reduce the number of constraints and variables in our ILP encoding.

\paragraph{Fusion of affine transformations and batch normalization}
In the inference phase, the parameters of a batch normalization layer are fixed, making it an affine transformation, e.g., $\yvec=BN(\xvec)=\gamma(\xvec-\mu_{\xvec})/\sqrt{\sigma_{x}^{2}+\epsilon}+\beta$. Two consecutive affine transformations can always be rewritten as a new single affine transformation. For example, $\yvec=BN(W\xvec+\bvec)$ can be regarded as a new affine transformation $\yvec=W'\xvec+\bvec'$. Therefore, a convolutional layer (or a linear layer) can be fused with a batch normalization layer to get a new convolutional layer (or linear layer), where the input tensor $\xvec$ remains unchanged but the weight and bias parameters are updated accordingly. In our ILP encoding, we also fuse such layers.  The encoding process is the same, but using the modified weights and biases.

\begin{figure}
\centering
\includegraphics[width=1\linewidth]{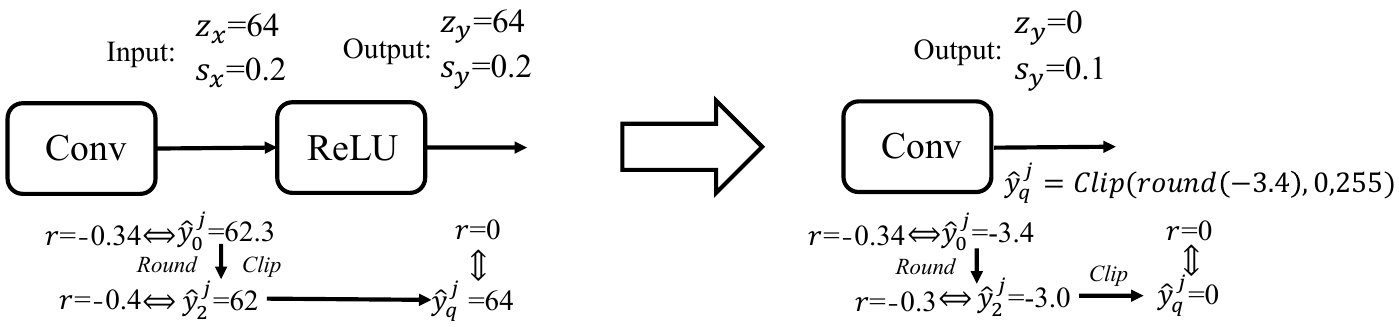}
\caption{Fusion of the convolutional layer and \ReLU.}
\label{fig:fusion}
\end{figure}

\paragraph{Fusion of affine transformations and ReLU} The \Clip operation has a similar function to \ReLU: they both limit the lower bound of the output. So, in the quantification process, these two operations can be fused together into one \Clip  operation. For example, in a quantized neural network, the convolutional and \ReLU layers can be merged together to form a $ConvReLU$ layer. For our encoding, we combine steps (iii) and (iv) to get a single \Clip operation:
\begin{equation}
  \text{(iii}\oplus \text{iv)} \ y_{q}^{j}:=Clip(\hat{y}_{1}^{j},lb',ub),
\end{equation}
where $lb'$ is the new lower bound. Fig. \ref{fig:fusion} shows an example. We can replace $lb$ and $\dot{\hat{y}}_{2}^{j}$ in constraint (\ref{clip_encoding}) with $lb'$ and $\dot{\hat{y}}_{q}^{j}$ to generate the encoding for (iii $\oplus$ iv).

\section{Interval Analysis}
In the second step of our verification framework, we compute the lower bound $lb(\dot{y})$ and upper bound $ub(\dot{y})$ for each variable $\dot{y}$. After obtaining the bounds for each variable, sometimes we can directly conclude that the property holds. Even if the bounds are not precise enough to prove the property, we can use them to simplify the ILP problem. In particular, the bounds may be able to show that some neurons are always active or always inactive. To compute the bounds, we use standard abstract interpretation techniques which use convex polyhedra to over-estimate the output interval of each node~\cite{WangPWYJ18}. We make two small contributions in this context that help support our goal of verifying QNNs.

First, we add support for the round operation, $\hat{y}_{1}^{j}:=Round(\hat{y}_{0}^{j})$. For this operation, the bounds on the output can easily be determined from the bounds on the input based on (\ref{round_encoding}), which can be rewritten as:

\begin{equation}
 \varepsilon-0.5+\hat{y}_{0}^{j}  \leq \hat{y}_{1}^{j} \leq 0.5+\hat{y}_{0}^{j}
\end{equation}

\begin{figure}[ht]
\centering
\includegraphics[width=0.8\linewidth]{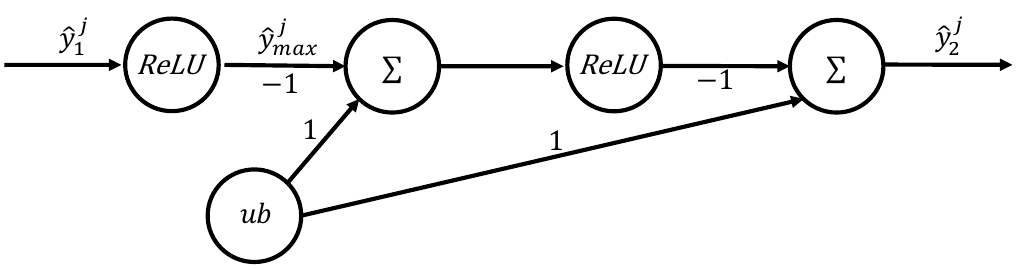}
\caption{Using two \ReLU{s} to simulate $\Clip(\cdot)$.}
\label{fig:d_relu}
\end{figure}

The other contribution is to support the clip operation, $\hat{y}_{2}^{j}:=Clip(\hat{y}_{1}^{j},lb,ub)$.  Our solution is to use \ReLU to simulate its function. The advantage of this method is that we can then leverage abstract interpretation techniques for \ReLU, which have been extensively studied and optimized~\cite{SinghGPV19,WuZ21}. The expression is:

\begin{equation}
\hat{y}_{max}^{j}:=\ReLU(\hat{y}_{1}^{j},lb), \hat{y}_{2}^{j}:=ub-\ReLU(ub-\hat{y}_{max}^{j})
\end{equation}
In other words, we can add the following structure  (Fig. \ref{fig:d_relu}) to the network and then use existing techniques to compute the bounds.

We implemented these two techniques in the Marabou neural network verification tool~\cite{KatzHIJLLSTWZDK19} which has support for abstract interpretation-based bound computation. This allows us to use Marabou to compute bounds for our QNNs.

\section{Gradient-based Heuristic Search}
Heuristic search can sometimes be very efficient at finding counterexamples to formal properties. We can thus use it as a complementary approach to abstract interpretation and exact verification. In particular, gradient-based heuristic search is an effective technique.  However, in popular implementations of quantization schemes (e.g., PyTorch), the gradients of QNNs are unavailable. One possible solution is to construct a new dummy neural network just for the purposes of gradient computation.  In the dummy network, we copy the structure and weights from the QNN but use the full floating-point representation. However, there is still an issue that must be addressed, which is that the \Round function is not differentiable. Indeed, it has derivative 0 at all points except where it is discontinuous.  However, if we look at the overall direction that the \Round function moves as we increase or decrease the input, it never goes too far away from $y=x$.  This suggests that using $y=x$ in place of $y=\Round(x)$ may be a good approximation.

For example, if we have $y=\Round(y_1)$ and $y_1=f(x)$, where $f$ is a differentiable function, we would like for $\partial y$/$\partial x$ to simply be computed as $\partial y_1$/$\partial x$.  To do so, the \Round operation can simply be omitted and we can just let $y=f(x)$ and get its gradient. However, this naive approach does not work when there are multiple layers with \Round operations. Suppose we have $y=\Round(y_3)$, $y_3=f_2(y_2)$, $y_2=\Round(y_1)$, $y_1=f(x)$, and we compute the gradient as follows:
\begin{equation*}
    \frac{\partial y}{\partial x}=\frac{\partial y_3}{\partial y_2}\frac{\partial y_1}{\partial x}.
\end{equation*}
We can see that the gradient value is related to the value of $y_2$ (the output of the \Round operation). But we have essentially dropped this value, so if we just remove all the \Round operations from the quantized neural network, this will cause an accumulation in the gradient error layer by layer.

To resolve this problem, we use a trick to rewrite the \Round operation in the dummy network so that both the output value and the gradient are available. For each $Round(\cdot)$ operation in our dummy network, we replace it with
\begin{equation}
    y=Round(x)+x-x.detach()
\end{equation}
where $x.detach()$ denotes the operation that returns a new tensor with the same value as $x$ but detached from the current computational graph. It is easy to see that the gradient of terms $\Round(x)$ and $x.detach()$ are $0$. So the value of the partial derivative of $y$ with respect to $x$ is the same as with the function $y=x$. Once we construct the dummy network, we can use a standard gradient-based attack to find counterexamples.  In our experiments, our implementation is similar to the PGD algorithm~\cite{MadryMSTV18}.

\section{Experiments}
We implemented a Python-based verification
tool called EQV\footnote{https://github.com/huangdiudiu/EQV}. We use Gurobi~\cite{Gurobi} as our backend ILP solver and the abstract interpretation is done by Marabou. The implementation of heuristic search and some results are described in the Appendix. 
The efficiency of our approach is evaluated on two well-known neural network architectures: fully connected neural networks (FC) and convolutional neural networks (CNN). We use the notation FCN-M to refer to a network consisting of N dense layers with M hidden units in each layer. For example, the structure of FC2-256 is 784 $\times$ 256$\times$ 256 $\times$ 10. CNN1 is a network with one convolutional layer of 4 channels, followed by one batch normalization layer, one max-pooling layer with a kernel size of 2, and a fully connected layer with 10 units. The convolutional layer has 4 $\times$ 4 filters and 2 $\times$ 2 strides with a padding of 1. CNN2 is identical to CNN1 except that its convolutional layer has 2 channels. All neural networks are trained on the MNIST dataset \cite{mnist} and quantized with PyTorch using its default static quantization scheme. Verification experiments are conducted on the test set. In particular, we verify the robustness of networks (using Equation~\eqref{rob}) with $r=4,8,12,16$. The experimental environment is a 20-core Intel(R) Xeon(R) E5-2640 v4 @ 2.40GHz CPU with 64GB of memory.

\begin{table}
\centering
\caption{Comparisons between QVIP and EQV.} 
\resizebox{0.35\textwidth}{!}{
\begin{tabular}{lccc}
\toprule
& & FC1-100 & FC2-100\\
\hline
\multirow{3}{*}{$r=4$} & QVIP & 11751.16 (30) &30000.00(100) \\
& ILP & 332.48 (0) & 4039.38 (7)\\
& ILP+In & \textbf{93.52 (0)} & 1581.41 (3)\\
& EQV & 104.09 (0) & \textbf{1465.46 (3)}\\

\hline
\multirow{3}{*}{$r=8$} & QVIP & 30000.00 (100)&30000.00(100)\\
& ILP & 2598.41 (7) & 25121.16 (77)\\
& ILP+In & 2729.48 (7) & 13854.10 (38) \\
& EQV & \textbf{1463.89 (2)} & \textbf{12064.63 (32)}\\
\hline
\multirow{3}{*}{$r=12$} & QVIP & 29512.08 (98)&30000.00(100)\\
& ILP & 10640.24 (31) & 29589.61 (98)\\
& ILP+In & 13467.12 (36) & 26069.18 (81)\\
& EQV & \textbf{7726.74 (17)} & \textbf{23427.27 (73)}\\
\hline
\multirow{3}{*}{$r=16$} & QVIP & 28945.63 (96)&30000.00(100)\\
& ILP & 18925.01 (57) & 29750.82 (98)\\
& ILP+In & 21172.09 (60) & 28902.99 (94) \\
& EQV & \textbf{6621.84 (15)} & \textbf{22724.67 (73)}\\
\bottomrule
\end{tabular}
\label{tab:qvse}
}
\end{table}

\begin{table*}
\centering
\caption{Total execution time(s) of different methods.}
\resizebox{0.75\textwidth}{!}{
\begin{tabular}{llrrrrr}
\toprule
& & FC2-256& FC2-512& FC3-100& CNN1& CNN2\\
\hline
\multirow{3}{*}{$r=4$}
& ILP &22111.22 (67) &29835.94 (97) & 29183.30 (95) & 28235.35 (94) & 28015.32 (93)\\
& ILP+In &4456.76 (11) &8013.00 (16) & 3347.88 (8) & 221.64 (0) & 1041.85 (2)\\
& EQV & \textbf{4019.10 (10)} & \textbf{7637.05 (15)} & \textbf{2421.66 (5}) & \textbf{205.48 (0)} & \textbf{723.13 (1)}\\
\hline
\multirow{3}{*}{$r=8$}
& ILP & 29991.71 (99) &30000.00 (100) & 30000.00 (100) & 25345.75 (84) & 27357.55 (90)\\
& ILP+In  & 22543.68 (62) & 29891.39 (89) &24330.21 (78) & 2503.19 (5) & 3715.99 (9)\\
& EQV & \textbf{19918.79 (55)} &\textbf{29179.91 (87)} & \textbf{22616.44 (72)} & \textbf{1850.32 (4)} & \textbf{1499.73 (3)}\\
\hline
\multirow{3}{*}{$r=12$}
& ILP  & 29660.15 (99)&30000.00 (100)& 29901.57 (99) & 21609.24 (70) & 27422.63 (90)\\
& ILP+In &29991.41 (99) &29930.07 (99) & 29559.77 (96) &5236.63 (11) & 13949.19 (38)\\
& EQV & \textbf{23938.37 (77)} &\textbf{29362.36 (92)} & \textbf{24137.56 (78)} & \textbf{4698.57 (11)} & \textbf{8333.80 (24)}\\
\hline
\multirow{3}{*}{$r=16$}
& ILP & 29991.33 (99) & 30000.00 (100) &29919.99 (99) & 15004.02 (46) & 28330.85 (93)\\
& ILP+In & 29767.65 (99) & 30000.00 (100) & 29928.29 (99) & 5084.85 (11)& 18977.19 (55)\\
& ILP+In+PGD & \textbf{12021.24 (50)} & \textbf{24647.06 (78)} & \textbf{16402.12 (54)} & \textbf{3897.42 (10)} & \textbf{6733.25(21)}\\
\bottomrule
\end{tabular}
}
\label{time}
\end{table*}
\subsubsection{Comparison}

To show the effectiveness of our strategies, three variants of our method: pure ILP, ILP with abstract interpretation (ILP+In), and EQV are involved in the experiments. We first compare with the SOTA QNN verification tool QVIP \cite{ZhangZCSZCS22} in terms of efficiency. What must be stated is that QVIP only supports a simplified quantization scheme and does not support the quantization scheme used in PyTorch. Although the network architecture used in the experiment is the same, the weights and computational processes of the network are not entirely identical. The size of FC2-100 has exceeded the maximum network size used in the experiment for QVIP. This experiment is only intended for reference. The comparison between our methods and EQV is shown in Table \ref{tab:qvse}. The number in ``()” indicates the number of timeouts (300s). We randomly select the 100 examples from the test set, and any instance that exceeds the timeout is recorded as 300 seconds.

 Table \ref{tab:qvse} demonstrates that EQV outperforms QVIP largely with less time and fewer timeouts. Our pure ILP method has achieved efficiency improvements of several tens of times compared to QVIP when the radius is small (e.g.$r=4,8$). Especially for FC-100, our methods improve efficiency by up to 78 times.
 Although the quantization scheme of the networks differs, comparisons between ILP and ILP+In against QVIP reveal that avoiding the use of piecewise constraints can significantly improve the verification efficiency.

\subsubsection{Performances on Larger NNs} Table \ref{time} shows the performance of our methods on some larger neural networks. From Table \ref{time}, we can see that: (1) abstract interpretation excels at handling cases with small radii, while heuristic search excels at finding counterexamples with large radii, confirming that these techniques are complementary; (2) in some cases, ILP+In is slower than ILP, indicating that abstract interpretation is not well-suited for these cases and simply adds overhead; however, when we add heuristic search, efficiency is improved, suggesting that heuristic search can sometimes compensate for the efficiency reduction caused by abstract interpretation; (3) for the same network, the efficiency of EQV decreases initially and then increases with the growth of the radius; the most challenging cases appear to occur near the maximum safe radius; (4) For some cases where the radius is less than 4, EQV is more than 100 $\times$ faster than the baseline.




\begin{figure}[ht]
\centering
\includegraphics[width=0.85\linewidth,height=0.65\linewidth]{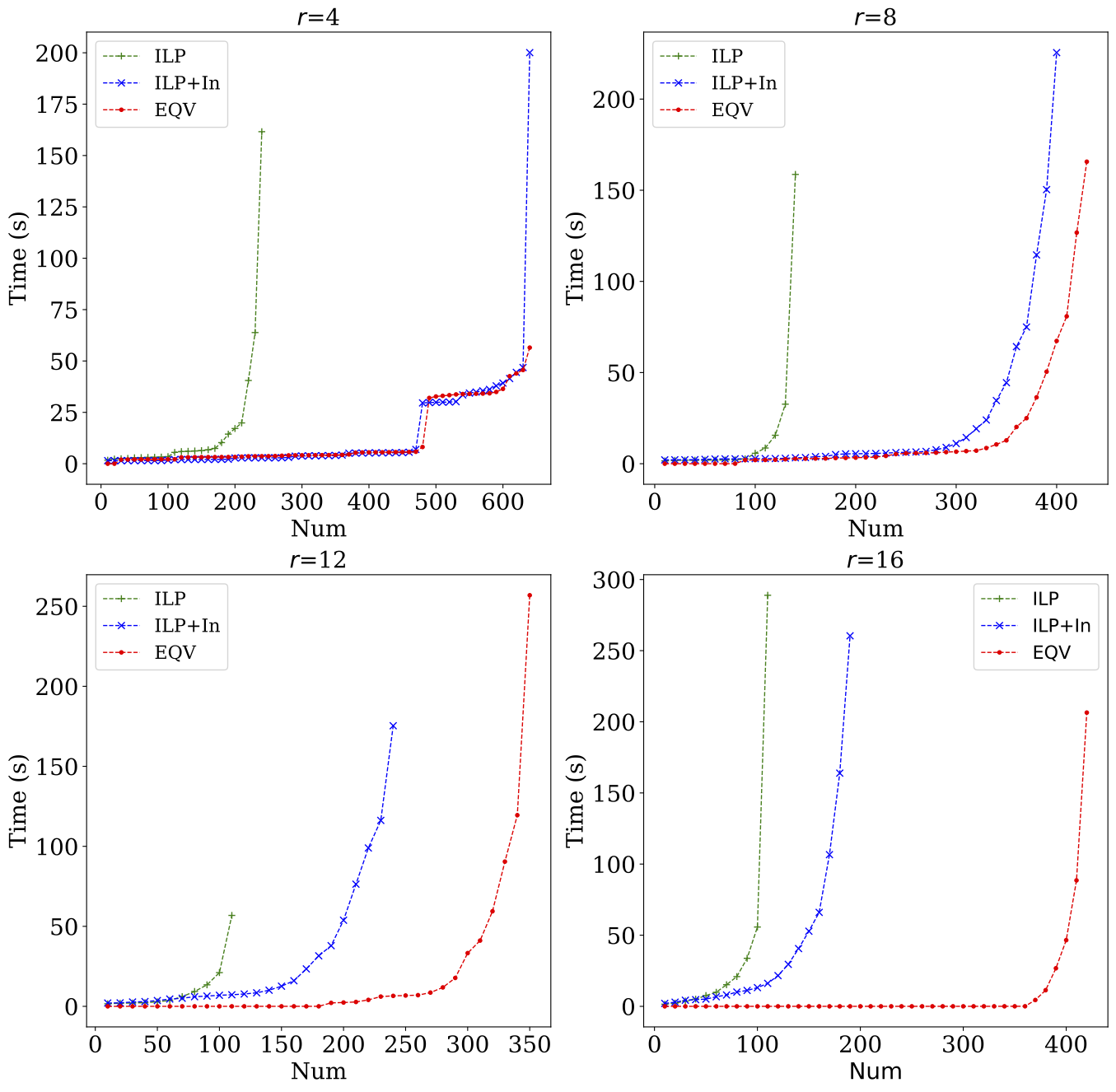}
\caption{Cactus plots on all the instances.}
\label{fig:time}
\end{figure}

Fig.~\ref{fig:time} provides a comprehensive assessment of the three methods across the entire set of solved instances, encompassing all seven networks. The $X$-axis represents the number of solved instances and the $Y$-axis represents the cumulative time needed to solve them. We can observe that EQV consistently demonstrates superior performance, and as the perturbation radius increases, its efficiency advantage over other methods becomes more pronounced.

\subsubsection{The Effectiveness of Different Parts}
\begin{figure}
\centering
\includegraphics[width=1\linewidth]{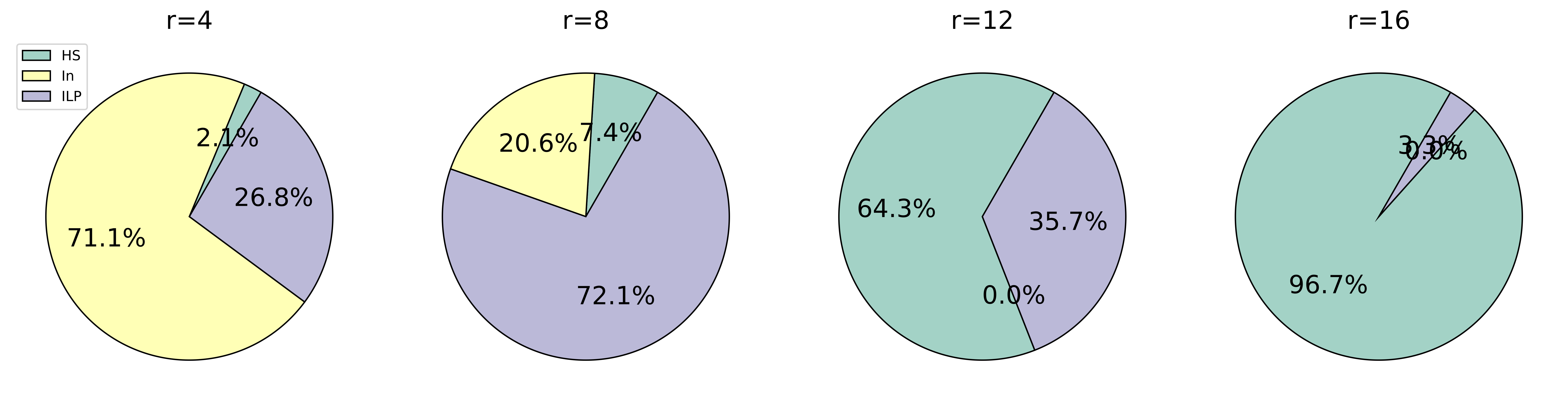}
\caption{The percentage of examples solved by each strategy (HS: heuristic search). }
\label{fig:PI}
\end{figure}

In the experiment, we recorded the contributions of each component of EQV. Taking FC-100 as an example, Figure \ref{fig:PI} shows the percentage of instances that were solved by each component. When the radius is relatively small, bound propagation primarily plays a major role in accelerating the solving of many instances; when the radius takes intermediate values, bound propagation can provide tighter bounds for ILP variables, thus expediting the solving process; when the radius is relatively large, bound propagation becomes ineffective, and heuristic search can compensate for efficiency loss by rapidly identifying counterexamples.

\subsubsection{Efficiency of Parallel Methods} 

\begin{figure}
\centering
\includegraphics[width=1\linewidth]{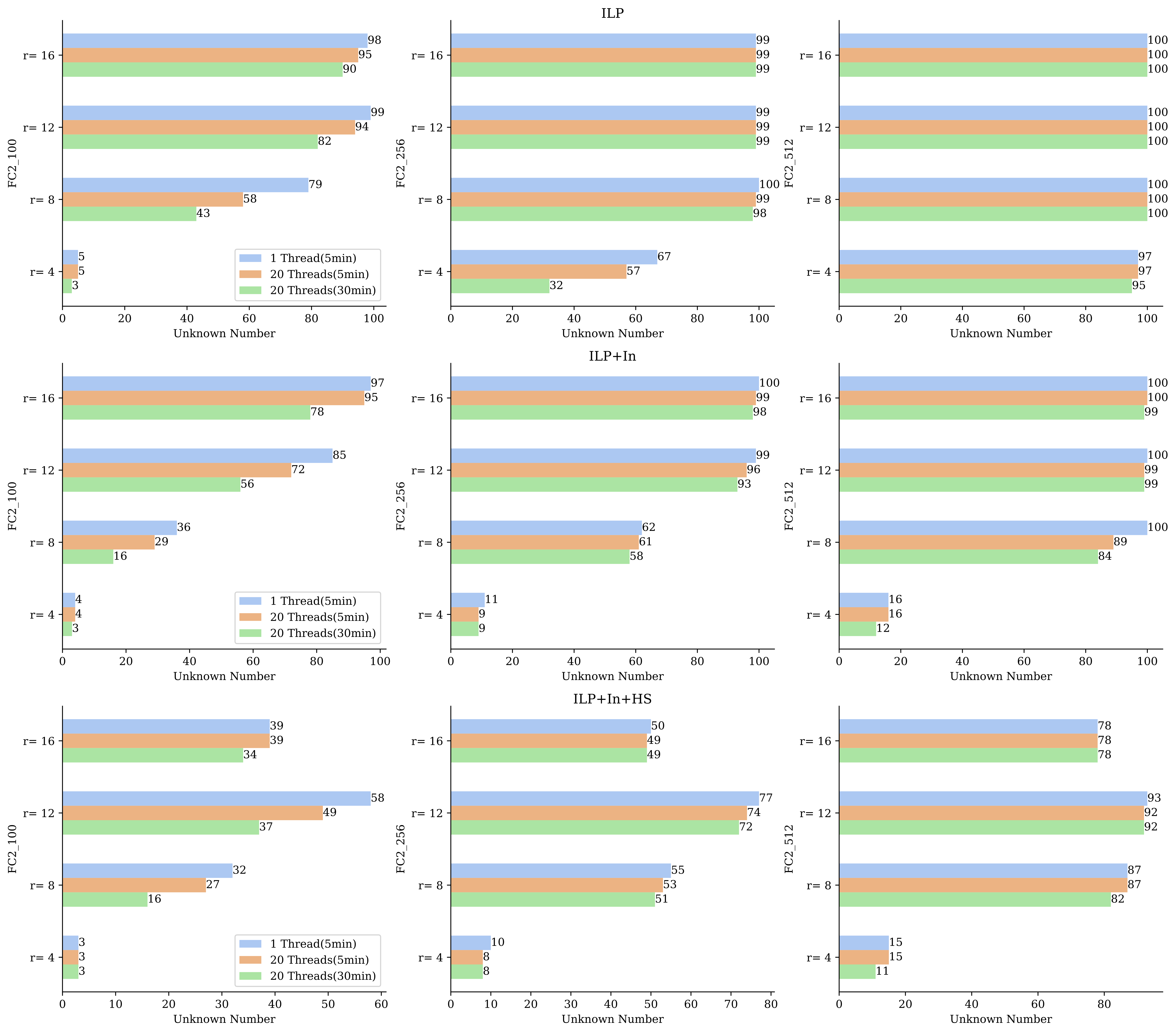}
\caption{The number of unknown instances under various configurations.}
\label{fig:parraltime}
\end{figure}

Based on Tables~\ref{time} and~\ref{tab:EQV}, it is clear that for larger networks and values of $r$, (e.g., for FC2-100, FC2-256 and FC2-512 when $r=4,8, 12, 16$), the problems become quite challenging for EQV. We did a preliminary investigation to determine whether parallel solving can help in these instances.  We ran the ILP and abstract interpretation methods with 20 parallel threads. Taking it a step further, to explore the impact of increased solving time on the success rate of verification, we also conducted experiments with a time limit of 30 minutes for each instance. The results are presented in Fig \ref{fig:parraltime}. Parallel algorithms do improve efficiency, but their improvements are less than 10\%. This indicates that a straightforward parallel approach has limited efficiency gains, and we need to explore parallel algorithms tailored for neural networks. There is still much room for improvement in parallel verification methods. We also observe that with extended time limits, the ILP method can sometimes solve nearly 50\% of the instances, but for EQV, providing more time yields only marginal improvements. This indicates that the complementary acceleration strategies employed within EQV have significantly leveraged the potential for accelerating the verification tool. Our acceleration strategy's effectiveness can be observed from Fig. \ref{fig:parraltime}, even surpassing the parallel acceleration for the basic method with 20 threads.


\subsubsection{Verification Results}

Table \ref{tab:EQV} shows the results given by EQV within 30 minutes. Acc is the accuracy of the QNN; ``Rob'' is the percentage of instances proven robust by EQV; ``Uns'' is the percentage shown to be unsafe by EQV; and ``Unk'' is the percentage that is unknown (i.e. timeouts). Notice that when $r = 4$, even for the largest network (FC2-512), our method provides an answer for nearly 88\% of the instances within the timeout.  This again suggests greater scalability than previous approaches.

\begin{table}
\centering
\caption{Results given by EQV (30 minutes).} 
\resizebox{0.48\textwidth}{!}{
\begin{tabular}{lcccccccc}
\toprule
& & FC1-100 & FC2-100& FC2-256& FC2-512& FC3-100& CNN1& CNN2\\
\hline
& Acc & 97.72\% & 97.96\% & 98.22\% & 98.33\% & 97.89\% & 95.36\% & 97.16\%\\
\hline
\multirow{3}{*}{$r=4$}
& Rob&96\%&95\%&90\%&88\%&93\%&90\%&89\%\\
& Uns&2\%&2\%&2\%&1\%&3\%&8\%&7\%\\
& Unk&0\%&3\%&8\%&11\%&4\%&0\%&0\%\\
\hline
\multirow{3}{*}{$r=8$}
& Rob&89\%&79\%&41\%&16\%&45\%&64\%&75\%\\
& Uns&9\%&5\%&8\%&2\%&6\%&33\%&20\%\\
& Unk&0\%&16\%&51\%&82\%&49\%&1\%&1\%\\
\hline
\multirow{3}{*}{$r=12$}
&Rob&62\%&36\%&6\%&1\%&12\%&31\%&43\%\\
& Uns&28\%&27\%&22\%&7\%&18\%&64\%&46\%\\
& Unk&8\%&37\%&72\%&92\%&70\%&3\%&7\%\\
\hline
\multirow{3}{*}{$r=16$}
&Rob&33\%&7\%&1\%&0\%&0\%&11\%&10\%\\
& Uns&57\%&59\%&50\%&22\%&46\%&87\%&78\%\\
& Unk&8\%&34\%&49\%&78\%&54\%&0\%&8\%\\
\bottomrule
\end{tabular}
\label{tab:EQV}
}
\end{table}

\begin{figure}
\centering
\includegraphics[width=0.95\linewidth, height=0.75\linewidth]{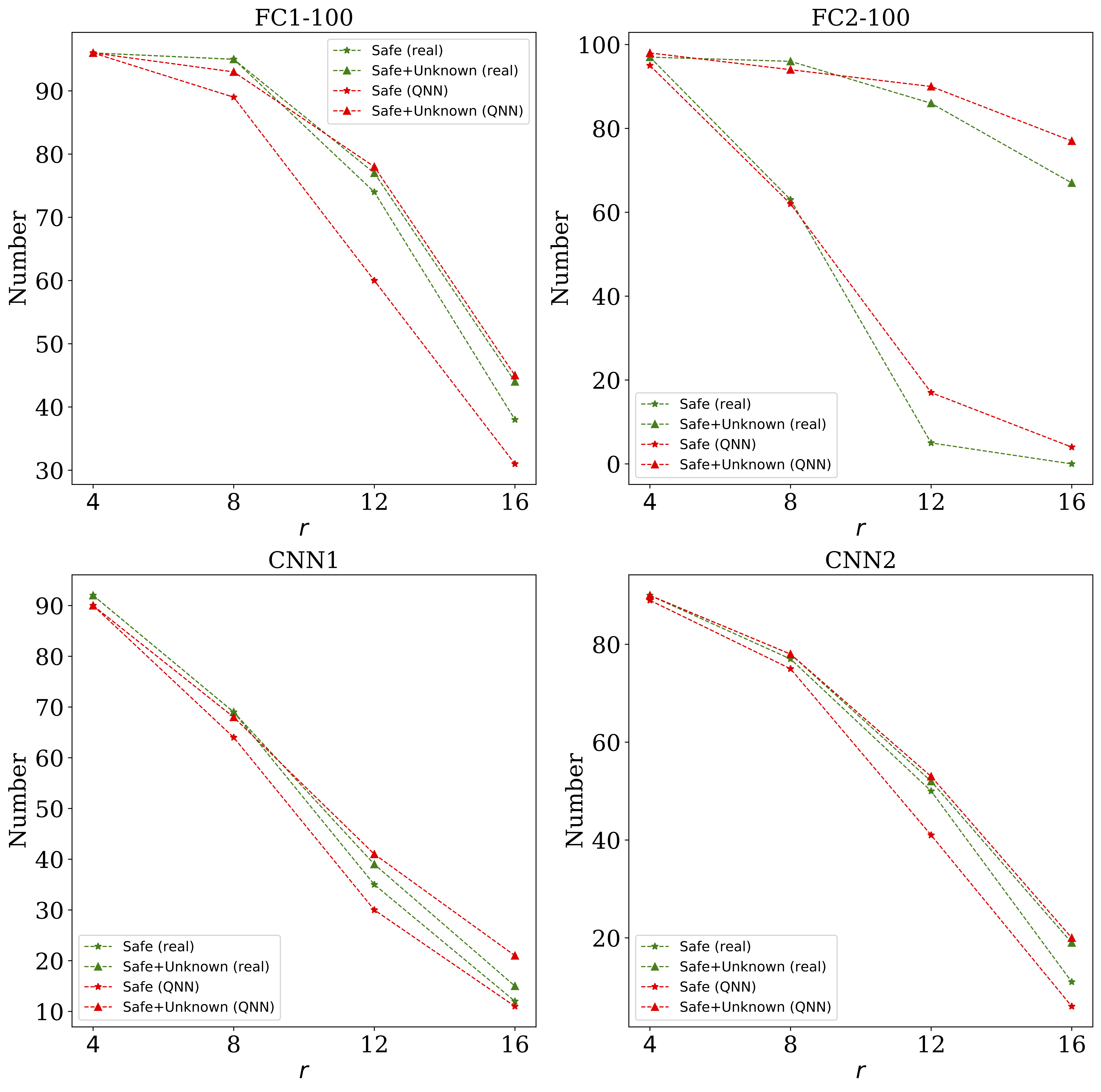}
\caption{Robustness curves of real-valued networks and their quantized version.}
\label{fig:quant-effect}
\end{figure}

\paragraph{Robustness Changes Caused by Quantization} To investigate the effect of quantization on the robustness of neural networks, we compare the robustness verification results of the original networks with those of the quantized neural networks. The experiments were conducted on 4 networks, namely FC1-100, FC2-100, CNN1 and CNN2. Fig.~\ref{fig:quant-effect} plots the percentage of instances shown to be safe at each radius. The curve of ``safe+unknown'' can be regarded as the success rate of resisting attacks. For QNN, the gap between the two curves ``safe+unknown'' and ``safe'' is larger than that of the real-valued network, indicating that the verification complexity of QNN might be greater than verifying a real-valued network. It is also interesting to observe that although quantization reduces the accuracy of a neural network, it does not always reduce the robustness of the network. From the point of view of resisting attack, when the radius is below a certain threshold, the real-valued networks exhibit better robustness, whereas when the radius exceeds a certain threshold, QNNs demonstrate superior robustness. 

\section{Conclusion}
In this work, we propose an efficient verification framework for QNNs that offers different trade-offs between scalability and precision. Our verification tool EQV is the first formal verification tool that precisely captures the quantization scheme used in popular deep learning frameworks. Although we focus on verifying adversarial robustness, our method could be generalized to verify other properties of QNNs. Experimental results show that EQV is more efficient and scalable than previously existing approaches. In future work, it would be interesting to formally analyze the difference or equivalence between the original networks and the quantized neural networks or to formally quantify the precision loss due to the quantization process.

\bibliography{aaai24}
\newpage

\section{Appendix}

\section{Details of PGD Settings}
This section is the details of PGD settings in Sec 7.

PGD is an attack method that is essentially
\textbf{p}rojected \textbf{g}radient \textbf{d}escent on the negative loss function.
\begin{equation}
    x^{t+1}=Proj(x^t+\alpha sgn(\nabla_{x^t}L(\theta, x^t, y)))
\end{equation}

In our experiments, $L$ is the cross-entropy. We set $\alpha=r/7$, where $r$ is the perturbation radius. For each instance, we iterate 7 rounds. And this search process is performed on the CPU.

To get the gradient of QNN, we reconstruct the computational graph of QNN with the float-point number and rewrite the $Round(\cdot)$ operation based on our method described in the main paper.

\section{Quantization Details}

Quantization refers to techniques for performing computations and storing tensors at lower bitwidths than floating point precision. A quantized model executes some or all of the operations on tensors with reduced precision rather than full precision (floating point) values. PyTorch supports multiple approaches to quantizing a deep learning model. In most cases, the model is trained in FP32 and then the model is converted to INT8. 

In our experiment, the quantization scheme we used is post-training static quantization. Post-training static quantization (PTQ static) quantizes the weights and activations of the model. It fuses activations into preceding layers where possible. It requires calibration with a representative dataset to determine optimal quantization parameters for activations. Post Training Static Quantization is typically used when both memory bandwidth and compute savings are important with CNNs being a typical use case.

\section{Experimental Results Supplement}
Table \ref{tab:resultchange} displays the number of instances where the safety of a neural network changes before and after quantization. We have documented this for perturbation radii of $r=4$, $8$, $12$, and $16$. For fully connected NN, as the radius increases, the proportion of areas experiencing changes in safety also increases. However, for convolutional neural networks, we observed a very interesting phenomenon: the proportion of the area where safety varies first increases and then decreases as the radius increases.
\begin{table}
\centering
\caption{Number of Instances where Safety Changed after Quantification.(NN$\rightarrow$QNN)} 
\resizebox{0.5\textwidth}{!}{
\begin{tabular}{lcccc}
\toprule
& & FC1-100 &CNN1&CNN2\\
\hline
\multirow{2}{*}{$r=4$} & SAFE $\rightarrow$ UNSAFE & 2 & 8 & 5\\
& UNSAFE $\rightarrow$ SAFE & 2 & 7 & 4\\
\hline
\multirow{2}{*}{$r=8$} & SAFE $\rightarrow$ UNSAFE & 9 & 23 & 16\\
& UNSAFE $\rightarrow$ SAFE & 3 & 19 & 14\\
\hline
\multirow{2}{*}{$r=12$} & SAFE $\rightarrow$ UNSAFE & 27 & 23 & 23\\
& UNSAFE $\rightarrow$ SAFE & 16 & 20 & 19\\
\hline
\multirow{2}{*}{$r=16$} & SAFE $\rightarrow$ UNSAFE & 24 & 8 & 10\\
& UNSAFE $\rightarrow$ SAFE & 22 & 8 & 10\\
\bottomrule
\end{tabular}
\label{tab:resultchange}
}
\end{table}

The scatter plot in Figure \ref{fig:scatter} shows the time required for verification of all instances which are classified into UNSAT(SAFE) and SAT(UNSAFE) categories. Each point represents an instance, with the $y$-axis displaying the time needed for its verification.

\begin{figure*}[ht]
\centering
\includegraphics[width=0.99\linewidth]{scatter.png}
\caption{temp.}
\label{fig:scatter}
\end{figure*}

\end{document}